\documentclass[11pt,a4paper]{article}
\usepackage{authblk}

\usepackage[hyperref]{acl2020}
\usepackage{times}
\usepackage{latexsym}
\usepackage{graphicx} 
\graphicspath{ {data/} }

\usepackage{url}

\usepackage{breakurl}
\usepackage{booktabs}

\usepackage{enumitem}

\usepackage{caption}

\usepackage{multibib}
\newcites{caselaw}{Case Law References}
\newcites{law}{Law Reference}

\newcommand{\captionlarge}[1]{\stepcounter{figure}\raisebox{-7pt}
  {\footnotesize Fig. \thefigure.\hspace{5pt} #1}}

\usepackage{textcomp} 

\usepackage{microtype}

\aclfinalcopy 
\def\aclpaperid{12} 

\title{Generating Intelligible \textit{Plumitifs} Descriptions: \\ Use Case Application with Ethical Considerations}

\author[1]{\textbf{David Beauchemin}\thanks{\hspace{4pt}Contributed equally to this work.}~\hspace{1.5pt}}
\author[1]{\textbf{Nicolas Garneau}\protect\footnotemark[1]~\hspace{1.5pt}}
\author[2]{\textbf{Eve Gaumond}\protect\footnotemark[1]~\hspace{1.5pt}}
\author[2]{\\\textbf{Pierre-Luc D\'eziel}}
\author[1]{\textbf{Richard Khoury}}
\author[1]{\textbf{Luc Lamontagne}}
\affil[1]{Department of Computer Science, Universit\'e Laval, Qu\'ebec, Canada}
\affil[2]{Faculty of Law, Universit\'e Laval, Qu\'ebec, Canada}
\affil[ ]{\texttt{\small{david.beauchemin.5@ulaval.ca,eve.gaumond@observatoire-ia.ulaval.ca}}\vspace{-4pt}}
\affil[ ]{\texttt{\small{\{nicolas.garneau,richard.khoury,luc.lamontagne\}@ift.ulaval.ca}}\vspace{-4pt}}
\affil[ ]{\texttt{\small{pierre-luc.deziel@fd.ulaval.ca}}}

\date{}

\begin{document}
\maketitle

\begin{abstract}
\textit{Plumitifs} (dockets) were initially a tool for law clerks.
Nowadays, they are used as summaries presenting all the steps of a judicial case. Information concerning parties' identity, jurisdiction in charge of administering the case, and some information relating to the nature and the course of the preceding are available through \textit{plumitifs}. They are publicly accessible but barely understandable; they are written using abbreviations and referring to provisions from the Criminal Code of Canada, which makes them hard to reason about. In this paper, we propose a simple yet efficient multi-source language generation architecture that leverages both the \textit{plumitif} and the Criminal Code's content to generate intelligible \textit{plumitifs} descriptions. It goes without saying that ethical considerations rise with these sensitive documents made readable and available at scale, legitimate concerns that we address in this paper.
This is, to the best of our knowledge, the first application of \textit{plumitifs} descriptions generation made available for French speakers along with an ethical discussion about the topic.

\end{abstract}

\section{Introduction}
\label{sec:intro}

The right to access judicial information is a fundamental component of Canadian democracy and its judicial process  \citecaselaw{VancouverSun, CbcvCanada}\footnote{Italic references refer to case laws.}.
This right has two main purposes. First, to enhance judicial accountability by providing opportunities to the public to scrutinize it and put forward criticisms of the judicial process \citecaselaw{SierraClub,CBCvNB}.
Second, it has an educational purpose: by accessing judicial information, people acquire a better understanding of the court process \citecaselaw{EdmontonJournal}.
Given these purposes, the necessity to provide access to judicial information in an intelligible form cannot be ignored. Indeed, getting a copy of a document is not enough; people have to understand its contents.
This is particularly crucial in a digital context since citizens face an overload of judicial information online \citep{Eltis2011}.
As a consequence, litigants have great difficulty in finding relevant information for their case online  \citep{BaharyDionne2019}.

Studies show that, in the province of Quebec, the \textit{plumitif} (a public register where one can find an official trace of all the actions taken by the courts) lacks intelligibility \citep{PromTep2019}.
Some users have called it “non-sense” for non-attorneys \citep{Parada2020}.
Yet, the \textit{plumitif} is necessary for every litigant as it provides information concerning the parties' identity, the jurisdiction responsible for administering cases, and information relating to the nature and the course of proceedings.
In this work, we aim at leveraging both information extraction and natural language generation to increase the intelligibility of excerpts of the Court of Quebec's \textit{plumitif} regarding criminal offenses under the Criminal Code of Canada (CCC). 

Improving the comprehension of textual legal documents has been the subject of several studies in the past.
For example, patent claims are long legal pieces of texts that contain complex sentences making it hard for a layperson to reason about.
\citet{Sheremetyeva2014AutomaticTS} framed this problem into an automatic text simplification task while \citet{Farzindar2004RsumDT} and \citet{Hachey2006ExtractiveSO} proposed extractive summarization techniques to make them easier to understand.
The \textit{plumitifs}, while also lying in the ``legal texts'' family, take a completely different form; they are not written in a valid grammatical form, and contain many abbreviations and references to the CCC.
This makes our use case application rather unique.




To handle this type of document, we have designed a \textit{description generation} pipeline, divided into three steps.
The first step consists of segmenting a \textit{plumitif} into different parts.
In the second step, we extract, for each part, the relevant information using a Named Entity Recognition (NER) model.
For the final step, we generate sentences from the data extracted by the NER model.
To this end, we use a template-filling approach to ensure there are no factual fallacies introduced in the generation, an essential concern in legal text generation.
Moreover, we use a statistical language model in a controlled setting to augment the generation with vital contextual information, namely texts from the CCC, making our approach a hybrid generation model. Our contributions, in this work, are twofold;
\setlist{nolistsep}
\begin{enumerate}[noitemsep]
    \item We propose a simple yet robust data-to-text multi-source textual generation pipeline to make \textit{plumitifs} easier to understand for the litigants (made available through a web application, see Appendix~\ref{sec:webapp});
    \item We bring a discussion on the ethical considerations about privacy and discrimination
    that such an application may cause.
\end{enumerate}

We further describe our architecture, related work and methodology in Section~\ref{sec:generation} and evaluate its generation capabilities in Section~\ref{sec:eval}.
We bring important ethical considerations in Section~\ref{sec:ethic}
and open the discussion for future work in Section~\ref{sec:conclusion}.





\section{Generating Intelligible \textit{Plumitifs} Summaries}
\label{sec:generation}

\textit{Plumitifs} are used as summaries presenting all the steps of a case heard by the court. In the context of criminal proceedings, they contain information about the plaintiff, the accused, different charges along with their associated penalty (if applicable).
We present a \textit{plumitif} example in Appendix~\ref{sec:trans_plumitif}, Figure~\ref{fig:plumitif}.
\textit{Plumitifs} are freely available in person at any courthouse and are also accessible on the \textit{Société québécoise d'information juridique} (SOQUIJ) website\footnote{\url{https://soquij.qc.ca/}} where they can be consulted for a fee.
In this section, we detail our proposed architecture, which is broken down into three steps; segmenting the \textit{plumitif} into parts, extracting relevant information from each part, and generating descriptions by also leveraging the CCC.
We illustrate the whole architecture in Appendix~\ref{sec:archi}, Figure~\ref{fig:plumitif_arch} and further detail each component in the following subsections.

\subsection{Segmenting the \textit{Plumitif}}
\label{subsec:segmenting}
We identify three parts in a \textit{plumitif}; the accused, the plaintiff, and the charges. Since the \textit{plumitif} structure is pretty regular, it allows us to identify each one using simple heuristics based on the presence of specific strings (e.g. ``ACC.'' for ``accused'') with 100\% accuracy.
Splitting into parts simplifies the NER step since these models typically use a narrow contextual window of a few tokens on either side to make their prediction.
It also provides more data points overall.

\subsection{Extracting Relevant Information}
\label{subsec:ie}

As mentioned in Section~\ref{sec:intro}, we frame the retrieval of the relevant entities in the \textit{plumitif} as an information extraction problem.
That is, given a raw part of the \textit{plumitif}, a NER model extracts entities from the text to fill in a normalized view.
We established nine types of entities that need to be extracted; \textit{Adresses} (Addresses), \textit{Accusations et spécifications d'accusations} (Charges and Charges Specifications), Dates, \textit{Décisions} (Decisions), \textit{Lois} (Laws), Accusations, \textit{Organisations} (Organizations), \textit{Personnes} (Persons),  \textit{Plaidoyer} (Pleas) and \textit{Peines} (Sentences).
For the rest of the paper, we will use the French entities within the French templates and rules, and the English entities otherwise (i.e. in the text).

We manually annotated 816 \textit{plumitifs} 
from eight districts over the last five years, to cover as much variety as possible.
These eight districts are the ones with the most cases for this date range.
We train a NER model on the annotated dataset, which achieves, on average, a F1-Score of 0.965, thanks to the regularity in the form the \textit{plumitifs} can take\footnote{Training details are available in Appendix~\ref{sec:training_ner}}.



Once the relevant information is extracted and normalized, we use it in the third step of the pipeline, which consists of a data-to-text generation model, described in the following subsection.

\subsection{Realisation of \textit{Plumitif} Summaries}

Even though statistical and deep Natural Language Generation (NLG) has seen tremendous breakthroughs in recent years \citep{Radford2018ImprovingLU, Radford2019LanguageMA, Brown2020LanguageMA},
we decide not to strictly rely on this kind of Transformer model~\citep{Vaswani2017AttentionIA} for our use case.
Several architectures \citep{Ziegler2019FineTuningLM, Keskar2019CTRLAC, Dathathri2020PlugAP} attempt to control the generation of such pre-trained models by using conditioning elements that propose a specific stylistic or emotion for example.
However, \citet{Brown2020LanguageMA} showed that one of the best neural language models to date (GPT-3) may generate non-factual utterances, often called hallucinations \citep{rohrbach-etal-2018-object, rebuffel2020parenting}, or even hide significant biases that may put the credibility of generation at stake.

Since we generate legal textual content that can be used in various sensitive applications (e.g. HR screening, \cite{Parada2020}), we can't afford to let a model ``statistically'' generate a non-factual decision (e.g. guilty but the accused is not) or a charge (e.g. something that the accused has not done). Thus, we prefer to sacrifice variability for control by using a template-filling approach. \citet{Puzikov2018E2ENC} showed that a template-based approach can be as good as a neural encoder-decoder model on generating restaurant descriptions from sets of key-value pairs.
\citet{Deemter2005RealVT} also argues that ``template-based approaches to the generation of language are not necessarily inferior to other [statistical] approaches as regards their maintainability, linguistic well-foundedness and quality of output''.
This approach has been shown recently to perform well in different areas like weather reports \cite{RamosSoto2015LinguisticDF}, financial analysis \cite{Nesterenko2016BuildingAS} and soccer game reports \citep{Lee2017PASSAD} where they are used in production.




\subsubsection{Template-Based, Data-to-Text Generation}


In the same way \citet{Deemter2005RealVT} did, we manually deduce 66 patterns from a subset of the \textit{plumitifs} to generate the description text using the extracted information from the model introduced in Section~\ref{subsec:ie}\footnote{We present a complete generation example in Appendix~\ref{app:example} based on the \textit{plumitif} presented in Figure~\ref{fig:plumitif}.}.
The generation rules (especially the sentence ones) have been written by a legal expert.
Following the example in Figure~\ref{fig:plumitif}, with the corresponding extracted information about the accused and a really simple yet efficient rule, we can generate texts about the accused and the plaintiff, as illustrated in Appendix~\ref{app:accused_example}.






In the next subsection, we present how we combine the information extracted from the \textit{plumitif} with a parsed version of the CCC \footnote{\url{https://laws-lois.justice.gc.ca/eng/acts/c-46/}} using a Masked Language Model.

\subsubsection{Leveraging the Criminal Code of Canada}
\label{subsub:mlm}

The \textit{Criminal Code of Canada} (CCC) is an act that contains most of the criminal law in Canada.
It contains around 1,500 provisions (referred to with numbers) where each of them comprises paragraphs and subparagraphs.
The \textit{plumitifs} refers to provisions from the law using only the provision numbers, which provides little to no context to the litigants.
Therefore, it is essential to extract the law's text from the Criminal Code when generating the \textit{plumitif}'s summary.
However, the CCC is only available in HTML or PDF format, making it hard to query it programmatically.
Thus, we parsed the HTML version into the JSON format, which allows us to easily query for different articles, paragraphs and subparagraphs~\footnote{We were able to properly extract the 1518 provisions publicly release the JSON version of the French CCC here: \url{https://bit.ly/3kiBdFd}}.

A \textit{plumitif} may contain several charges.
Each charge may refer to one or two provisions from the law.
The first provision is most likely referring to the description of the law, where the title briefly summarizes the description.
The second provision (if any) is usually there to specify the charge~\footnote{In this work, we do not leverage the second provision.}.

Given the following template (see Appendix~\ref{sec:stiching_trans} for a translated version);

\begin{flushleft}
    \small
    \texttt{\textbf{<Accusé>} est accusé \textbf{<Article>}.}
\end{flushleft}

we wish to insert the provision title syntactically.
To this end, we propose to ``stitch'' the two pieces of the template using a Masked Language Model.
We use the French pre-trained version of BERT~\citep{Devlin2019BERTPO}, CamemBERT~\cite{martin2020camembert}, which has been trained on the French subset of OSCAR~\citep{ortiz-suarez-etal-2020-monolingual}, a huge multilingual corpus obtained by language classification and filtering of the Common Crawl corpus.

One of BERT's abilities is to predict randomly masked tokens in a sentence, usually referred to as a \textit{Cloze} task in the literature \citep{Taylor1953ClozePA}.
We specifically leverage this ability to our benefit, and
let CamemBERT predict the proper preposition that should be inserted between the template and the charge's title (\textit{défaut de se conformer à une ordonnance} here).
The realisation of the previous template would then look like the following (Appendix~\ref{sec:stiching_trans});

\begin{flushleft}
    \small
    \texttt{\textbf{John Doe} est accusé \textbf{pour} défaut de se conformer à une ordonnance.}
\end{flushleft}

Using the 134 unique charges titles included in our dataset, we find that CamemBERT can predict the right preposition 84\% of the time.





\subsubsection{Pleas, Decisions and Sentences}
\label{subsubsec:sentences_gen}
The generation of the pleas and decision text is simple since there are only a few possible situations, using 14 generation rules out of the 66 deduced.
For the first, it is either guilty or not guilty.
For the second, it is guilty, not guilty, or ten other technical situations such as ``arret'' (i.e. case where the court orders a stay of proceedings).
In both cases, the mapping between the pleas and decision is one-to-one with the associated generated text (i.e. a guilty decision can generate only one text).
We illustrate this case in Appendix~\ref{app:decision_example}.

On the other hand, generating Sentences is more complex.
In our set of 66 deduced generation rules, 50 are used to generate the Sentences.
This complexity is mostly due to the occurrences of different convictions in one Sentence, meaning that the mapping is one-to-many (i.e. a Sentence can have an unknown number of convictions).
Given the Sentence's extracted convictions, we order them by types (i.e. the penalty inflicted of, fines and fees, community work, other convictions, probation and surcharge) and fill-in an ``on-the-fly merged generation template'' given the list of convictions.
It is important to note that generation rules are not applied ``in cascade'' i.e. for a given list of convictions, there is one possible generation template.
We illustrate the generation of the first Sentence's section in Appendix~\ref{app:sentence_example}.

\section{Evaluating the Realisation of the Summaries}
\label{sec:eval}

Since our generation model mostly relies on rules, it is straightforward to evaluate its performance; we first need to make sure all the relevant information is fully extracted (NER step) and that it properly fills in the corresponding template (generation step).
We thus quantify our model's performance in terms of ``Error Rate'' where a generation error is the lack of realizing a specific part (accused, plaintiff or list of charges paragraphs), instead of evaluating the textual generation.
The counts are computed per text.
Errors are split into two categories; Extraction-based Errors (EE) and Generation-based Errors (GE).
For clarity, we display the Errors Rates by districts in Table~\ref{tab:error_rates}.

In most cases, we find that a wrong extraction of the Plaintiff (due to the NER model) causes EE.
We can see that Granby and Sherbrooke have the highest EE rate; this is mostly due to the many different values an Organisation can take in these districts~\footnote{This corroborates with the results of the NER model for the entity \textbf{Organisation}, in Section~\ref{subsec:ie}}.


GE are mainly due to edge cases found in \textit{plumitifs} which our rules do not cover.
As we can see from the GE Rates in Table~\ref{tab:error_rates}, our generation rules commit most errors on the Montréal, Sherbrooke and Gatineau districts.
This is due to the numerous and diverse convictions these \textit{plumitifs} hold.
For example, a particular combination of convictions may not be associated to any generation rule.
We illustrate this problem with an example in Figure~\ref{fig:sentence_example}, where the Sentence comprises multiple convictions and are essentially edge cases about the duration.

\begin{table}[h!]
    \centering
    \begin{tabular}{r||cc|c}
        \textbf{District} & \textbf{EE} & \textbf{GE} & \textbf{\textit{Plumitifs}}\\
        \hline
        Chicoutimi & 0.0\% & 0.0\% & 9\\
        Gatineau & 6.7\% & 6.7\% & 15\\
        Granby & 33.3\% & 5.6\% & 18\\
        Longueuil & 5.9\% & 0,0\% & 17\\
        Montréal & 13.8\% & 9.2\% & 65\\
        Québec & 0.0\% & 0.0\% & 18\\
        Sherbrooke & 25.0\% & 8.3\% & 12\\
        Trois-Rivières & 15.4\% & 0.0\% & 13\\
        \hline
        \textbf{Average} & \textbf{13\%} & \textbf{5\%}\\
    \end{tabular}
    \caption{Error rates of the Extraction (EE) and Generation (GE) errors for each district.}
    \label{tab:error_rates}
\end{table}

\begin{figure}[h!]
\small
    \flushleft
    PROBATION DE  2 ANS SURV.\\
    PROBATION DPAC:8.5MS/EMPR:6.5M\\                   
    TC 75 HS DEL 12 MS/SUIVI PROB 1 1/2 AN
    \caption{Example of a complex Sentence containing a edge case about the duration of the different convictions. For this specific example, our model failed to generate a meaningful piece of text.}
    \label{fig:sentence_example}
\end{figure}

This highlights the need to have a better model at parsing \textit{and} generating Sentences' paragraphs.
Using a generative, sequence-to-sequence model, such as the one proposed by \cite{Bahdanau2015NeuralMT} may be a better option, but we leave this study as future work.
All in all, our model achieves low Error Rates (13\% EE and 5\% GE on average), allowing simple yet accurate textual generation of intelligible plumitifs.
While these results are interesting, it raises some ethical concerns, that we discuss in the next section.

\section{Ethical Considerations}
\label{sec:ethic}

There is some ethical considerations regarding our dataset’s privacy that ought to be addressed. \textit{Plumitifs} contain sensitive information such as the names, dates of birth, addresses and criminal backgrounds of accused people. The identity of judges, plaintiffs, clerks, and attorneys taking part in a criminal case are also found in the \textit{plumitifs}. As explained in Section~\ref{sec:intro}, all of this information must be publicly accessible. As long as this data is protected by practical obscurity~\footnote{A term broadly used to explain that documents might be accessible to all in principle, but that the access is hindered by some obstacles such as fees to consult a document or the need to go physically to a location - as is the case for the \textit{plumitif}.}, the actual risks from public access of this information are limited \cite{Vermeys2016}. 

However, if this data was to be released in bulk to the scientific community, it would not be ``scattered [$\ldots$] bits of information'' \citecaselaw{USDOJ} that require time and resources to retrieve anymore. Information could be easily searched, aggregated or combined with information from other public sources.  This poses a risk to the privacy of judicial stakeholders. 

In this subsection, we explain why we decided not to release our data set publicly (raw or synthesized). To put it in straightforward terms:  information collected in public records should not be "up for grabs". Its use can result in privacy violations.  This is especially true in the digital context where aggregation, linkage and analytics are made easier \citep{NissembaumMartin}.  
There are several examples of privacy violations that occurred due to the malicious use of judicial information that was publicly accessible. For instance, more than 270 cases of identity theft have been linked to a  security lapse in an American Municipal Court's website. \cite{BaileyBurkell2017}.
The Office of the Privacy Commissioner of Canada had to intervene to end an extortion scheme relying on data available from the Canadian Legal Information Institute and SOQUIJ’s websites \citecaselaw{Globe24h}. United State's ``Public Access to Court Electronic Records'' system made the identity of some cooperating defendants and undercover agents publicly available, which contributed to the intimidation and harassment of witnesses in order to discourage them from testifying \cite{Eltis2011}.
There have also been some documented cases of discrimination in the context of employment \cite{SoloveAggregation} and housing \citecaselaw{Purewal} caused by judicial information available online.
Moreover, academics have expressed significant concerns about the secondary use of judicial information for marketing purposes.

This is now prohibited by the Personal Information Protection and Electronic Documents Act, \citep{OPC2014}, but \cite{BaileyBurkell2017} argues that this regulatory framework is not sufficient to prevent inappropriate uses of judicial data. Our team is currently working to develop a framework for the management of personal information contained in digital court records. However, for the moment, since the law provides no satisfactory solution, we chose not to release the dataset used to train our algorithm. 

\section{Conclusion and Future Work}
\label{sec:conclusion}

In this paper, we introduce a simple yet effective multi-source architecture able to generate digestible \textit{plumitifs} for Canadian citizens.
We also show that we are in a position to easily divulge who has been accused of what and the outcome of it, which raises some important ethical concerns.
In the future, we plan to explore statistical natural language generation further by using case law, provide more diverse \textit{plumitifs} descriptions and improve the generation of Sentences.
Finally, we hope that our application will provide better insights to the community and give the right direction for the next applications of not only NLG, but Machine Learning in general, in the field of law.

\section{Acknowledgements}
We thank the reviewers for their insightful comments on our manuscript.
This research was enabled in part by the support of the Natural Sciences and Engineering Research Council of Canada (NSERC) and the Social Sciences and Humanities Research Council (SSHR). Also, this research was made possible with the help of our partner, \mbox{SOQUIJ}.

\bibliography{acl2020}
\bibliographystyle{acl_natbib}

\bibliographystylecaselaw{acl_natbib}
\bibliographycaselaw{acl2020}


\clearpage

\appendix

\section{\textit{Plumitif} Example}
\label{sec:trans_plumitif}

\begin{figure}[h]
    \centering
    \includegraphics[width=0.4\textwidth]{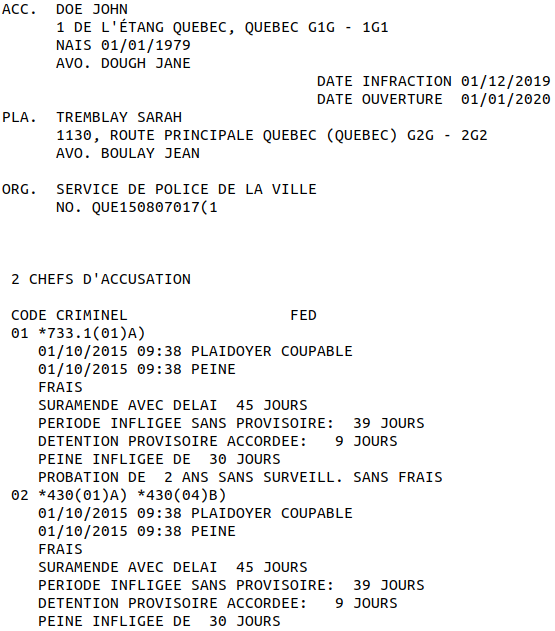}
    \caption{\textit{Plumitif} example illustrating the accused and plaintiff personal information along with charges and associated pleas, decisions and penalty. Names, dates and addresses have been edited to preserve privacy.}
    \label{fig:plumitif}
\end{figure}

\begin{figure}[h!]
    \centering
    \includegraphics[width=0.45\textwidth]{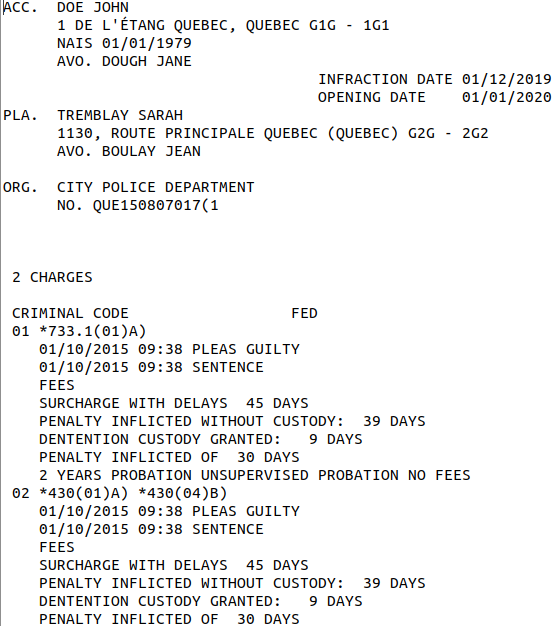}
    \caption{The translated version of the \textit{plumitif} example presented in Figure~\ref{fig:plumitif}.}
\end{figure}

\section{Architecture}
\label{sec:archi}

\begin{figure*}[h!]
    \centering
    \includegraphics[width=\textwidth]{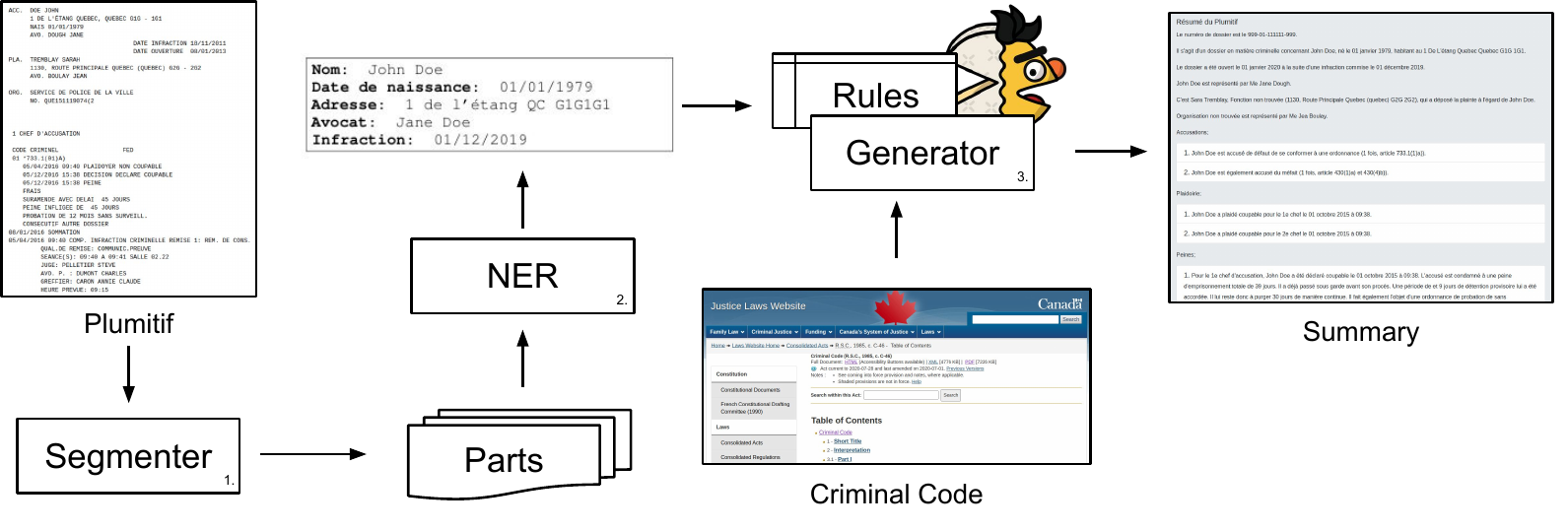}
    \caption{Overview of our three steps architecture that generates intelligible \textit{plumitifs} summaries. A \textit{plumitif} is first segmented into sections (1), which are then sent to the Named Entity Recognizer (NER) model, normalizing the relevant information (2). The extracted information, combined with the CCC, is used to generate a summary by leveraging both simple generation rules and a statistical Masked Language Model (3).}
    \label{fig:plumitif_arch}
\end{figure*}

\section{NER Training Details}
\label{sec:training_ner}

We split each district's \textit{plumitifs} into a training and testing set of roughly 80\%--20\% examples for evaluation purpose.
The numbers of \textit{plumitifs} per district are shown in Table~\ref{tab:districts_stats}, and occurrences of the different entities are displayed in Table~\ref{tab:ner_results}\footnote{We discuss in Section~\ref{sec:ethic} why we choose not to release the dataset.} (Occ.).

\begin{table}[h]
    \centering
    \begin{tabular}{r||ccc}
        \textbf{District} & \textbf{Train} & \textbf{Test} & \textbf{Total} \\
        \hline
        Chicoutimi & 35 & 9 & 44\\
        Gatineau & 59 & 13 & 72 \\
        Granby & 71 & 18 & 89\\
        Longueuil & 65 & 17 & 82\\
        Montréal & 253 & 65 & 318\\
        Québec & 72 & 18 & 90\\
        Sherbrooke & 48 & 12 & 60\\
        Trois-Rivières & 48 & 13 & 61\\
        \hline
        \textbf{Total} & \textbf{651} & \textbf{165} & \textbf{816}
    \end{tabular}
    \caption{Number of \textit{plumitifs} that we annotated, separated by districts and split in train and test sets.}
    \label{tab:districts_stats}
\end{table}

\begin{table}[h]
    \centering
      \resizebox{0.49\textwidth}{!}{%
    \begin{tabular}{r||c|c|c|c}
        \textbf{Entity} & \textbf{Precision} & \textbf{Recall} & \textbf{F1-Score} & \textbf{Occ.} \\
        \hline
        \textbf{Address} & 0.997 & 0.991 & 0.994 & 1649\\
        \textbf{Charge} & 0.980 & 0.982 & 0.981 & 3984\\
        \textbf{Date} & 0.995 & 0.998 & 0.996 & 8499\\
        \textbf{Decision} & 0.991 & 0.988 & 0.990 & 2374\\
        \textbf{Law} & 0.904 & 0.904 & 0.904 & 886\\
        \textbf{Organisation} & 0.905 & 0.910 & 0.910 & 845\\
        \textbf{Person} & 0.986 & 0.986 & 0.986 & 3146\\
        \textbf{Pleas} & 1.000 & 1.000 & 1.000 & 1956\\
        \textbf{Sentence} & 0.916 & 0.924 & 0.920 & 1609\\
        \hline
        \textbf{Average} & \textbf{0.964} & \textbf{0.965} & \textbf{0.965} & - 
    \end{tabular}%
		}
    \caption{Results of the NER model on the test set. Metrics are on the "entities" level, which means for a multi-token entity, if only one token is missing in the prediction, the prediction is wrong. Occurrences (Occ.) of the entities are on the full annotated dataset.}
    \label{tab:ner_results}
\end{table}

For the NER model, we use the sequence-to-sequence neural network model provided in the SpaCy library~\citep{spacy2}, which is based on a deep convolution neural network.
To train the model, we split every \textit{plumitif} into parts as described in Section~\ref{subsec:segmenting}; then, the model predicts the entities for each part separately instead of over the whole \textit{plumitifs}.
The results for the evaluation set can be seen in Table~\ref{tab:ner_results}.

\section{Accused Generation Example}
\label{app:accused_example}

Given the extracted information about the accused in the following form (we first present the original version in French followed by the English one);

\vspace{7.5pt}
\noindent\fbox{%
\small
    \parbox{0.45\textwidth}{%
    \texttt{\textbf{Nom}: John Doe}\\
    \texttt{\textbf{Date de naissance}: 01/01/1979}\\
    \texttt{\textbf{Adresse}: 1 de l'étang QC G1G1G1}\\
    \texttt{\textbf{Avocat}: Jane Doe}\\
    \texttt{\textbf{Infraction}: 01/12/2019}
    }%
}
\setlength{\fboxrule}{0.8pt}\noindent\fbox{%
\small
    \parbox{0.45\textwidth}{%
    \texttt{\textbf{Name}: John Doe}\\
    \texttt{\textbf{Date of Birth}: 01/01/1979}\\
    \texttt{\textbf{Address}: 1 de l'étang QC G1G1G1}\\
    \texttt{\textbf{Lawyer}: Jane Doe}\\
    \texttt{\textbf{Infraction}: 01/12/2019}
    }%
}

\vspace{7.5pt}
and given the following template;
\begin{center}
    \small
    \texttt{\textbf{<Accusé>}, né le \textbf{<Date de naissance>} habitant au \textbf{<Adresse>}, a commis une infraction le \textbf{<Date d'infraction>}. L'accusé est représenté par Me \textbf{<Avocat>}.}
\end{center}
\begin{center}
    \small
    \flushleft
    \texttt{\textbf{<Accused>}, born on \textbf{<Date of birth>} and living on \textbf{<Address>}, commited an infraction \textbf{<Infraction date>}. The accuse is represented by \textbf{<Lawyer>}.}
\end{center}

we can generate the following paragraph~\footnote{We will henceforth write templates filled with dynamic values in bold, in order to reduce repetition.};

\begin{center}
\small
    \texttt{\textbf{John Doe}, né le \textbf{1\textsuperscript{er} janvier 1979} habitant au \textbf{1 de l'étang QC G1G1G1}, a commis une infraction le \textbf{1\textsuperscript{er} décembre 2019}. L'accusé est représenté par Me \textbf{Jane Doe}.}
\end{center}
\begin{center}
\small
\flushleft
    \texttt{\textbf{John Doe}, born on \textbf{January 1st 1979} and living on \textbf{1 de l'étang QC G1G1G1}, committed an infraction \textbf{December 1st 2019} and is represented by \textbf{Jane Doe}.}
\end{center}

\section{Decision Generation Example}
\label{app:decision_example}
For example, given the following two decisions;

\setlength{\fboxrule}{0.2pt}\noindent\fbox{%
\small
    \parbox{0.45\textwidth}{%
    \texttt{\textbf{Décision 1}: arret}\\
    \texttt{\textbf{Date Décision 1}: 01/01/2020}\\
    \texttt{\textbf{Décision 2}: n-resp.tr.ment}\\
    \texttt{\textbf{Date Décision 2}: 01/01/2020}
    }%
}
\setlength{\fboxrule}{0.8pt}\noindent\fbox{%
\small
    \parbox{0.45\textwidth}{%
    \texttt{\textbf{Decision 1}: stop}\\
    \texttt{\textbf{Date Decision 1}: 01/01/2020}\\
    \texttt{\textbf{Decision 2}: n-lia.tr.ment}\\
    \texttt{\textbf{Date Decision 2}: 01/01/2020}
    }%
}
\vspace{7.5pt}

we can fill in the corresponding template and generate the following paragraphs for both decisions;


\begin{flushleft}
    \small
    \texttt{Pour le \textbf{1\textsuperscript{er}} chef d'accusation, le Tribunal prononce \textbf{un arrêt de procédure} le \textbf{1\textsuperscript{er} janvier 2020}. 
    Pour le \textbf{2\textsuperscript{e}} chef d'accusation, le Tribunal prononce \textbf{un verdict de non-responsabilité criminelle pour cause de troubles mentaux} le \textbf{1\textsuperscript{er} janvier 2020}.}
\end{flushleft}

we can generate the following sentence for both the decisions;
\begin{center}
    \small
    \flushleft
    \texttt{For the \textbf{1st} charge, the Court pronounces \textbf{a procedural judgment} on \textbf{January 1, 2020}. 
    For the \textbf{2nd} charge, the Court pronounces \textbf{a verdict of not criminally responsible on account of mental disorder} on \textbf{January 1, 2020}.}
\end{center}

\section{Sentence Generation Example}
\label{app:sentence_example}
The extracted information about the first Sentence~\footnote{We use 31 rules to extract the information from the Sentences. We discuss in Section~\ref{sec:eval}, a possible solution to circumvent the problems that this actual method introduces.} is then in the following form;

\setlength{\fboxrule}{0.2pt}\noindent\fbox{%
\small
    \parbox{0.45\textwidth}{%
    \texttt{\textbf{1}: Suramende \textbf{Délai}: 45 jours}\\
    \texttt{\textbf{2.1}: Provisoire \textbf{Durée}: 39 jours}\\
    \texttt{\textbf{2.2}: Accordée \textbf{Durée}: 9 jours}\\
    \texttt{\textbf{2.3}: Infligée \textbf{Durée}: 30 jours}\\
    \texttt{\textbf{3}: Probation \textbf{Durée}: 2 ans \textbf{Type}: sans surveillance}
    }%
}
\setlength{\fboxrule}{0.8pt}\noindent\fbox{%
\small
    \parbox{0.45\textwidth}{%
    \texttt{\textbf{1}: Surcharge \textbf{Delay}: 45 days}\\
    \texttt{\textbf{2.1}: Custody \textbf{Duration}: 39 days}\\
    \texttt{\textbf{2.2}: Pre-trial \textbf{Duration}: 9 days}\\
    \texttt{\textbf{2.3}: Inflicted \textbf{Duration}: 30 days}\\
    \texttt{\textbf{3}: Probation \textbf{Duration}: 2 years \textbf{Type}: unsupervised}
    }%
}
\vspace{7.5pt}


and given the corresponding filled-in template we generate the following Sentence paragraph;
\begin{flushleft}
    \small
    \texttt{L'accusé est condamné à une peine d'emprisonnement totale de \textbf{30 jours}. Il a déjà passé \textbf{39 jours} sous garde avant son procès. Une période de \textbf{9 jours} de détention provisoire lui a été accordée. Il lui reste donc à purger \textbf{21 jours} de manière continue. 
    Il fait également l'objet d'une ordonnance de probation de \textbf{2 ans} \textbf{sans surveillance}.
    Le paiement des frais de justice et de la suramende compensatoire qui sera versé dans un fond pour venir en aide aux victimes d'actes criminel doit être payé dans un délais de \textbf{45 jours}.}
\end{flushleft}


\begin{center}
    \small
    \flushleft
    \texttt{The accused is sentenced to a total imprisonment of \textbf{30 days}. He has already spent \textbf{39 days} in custody before his trial. He was granted a period o \textbf{9 days} in pre-trial detention. He therefore has to purge \textbf{21 days} continuously. 
    He is also subject to a probation order of \textbf{2 years} \textbf{unsupervised}. 
    The payment of court costs and the victim fine surcharge that will be paid into a fund to help victims of crime must be paid within \textbf{45 days}.}
\end{center}

\section{``Stitching'' Charges Translation}
\label{sec:stiching_trans}

Given the following template;
\begin{center}
    \small
    \flushleft
    \texttt{\textbf{<Accused>} is accused \textbf{<Charge>}.}
\end{center}
we wish to insert the charge title syntactically.

Given the updated template;

\begin{center}
    \small
    \flushleft
    \texttt{\textbf{<Accused>} is accused \textbf{<mask>} failure to comply with probation order.}
\end{center}

The realisation of the previous template would then look like the following;

\begin{center}
    \small
    \flushleft
    \texttt{\textbf{John Doe} is accused \textbf{for} failure to comply with probation order.}
\end{center}

\clearpage
\section{Complete Generation Example}
\label{app:example}
\begin{figure}[h!]
    \centering
    \captionsetup{width=\textwidth}
    \includegraphics[width=\textwidth]{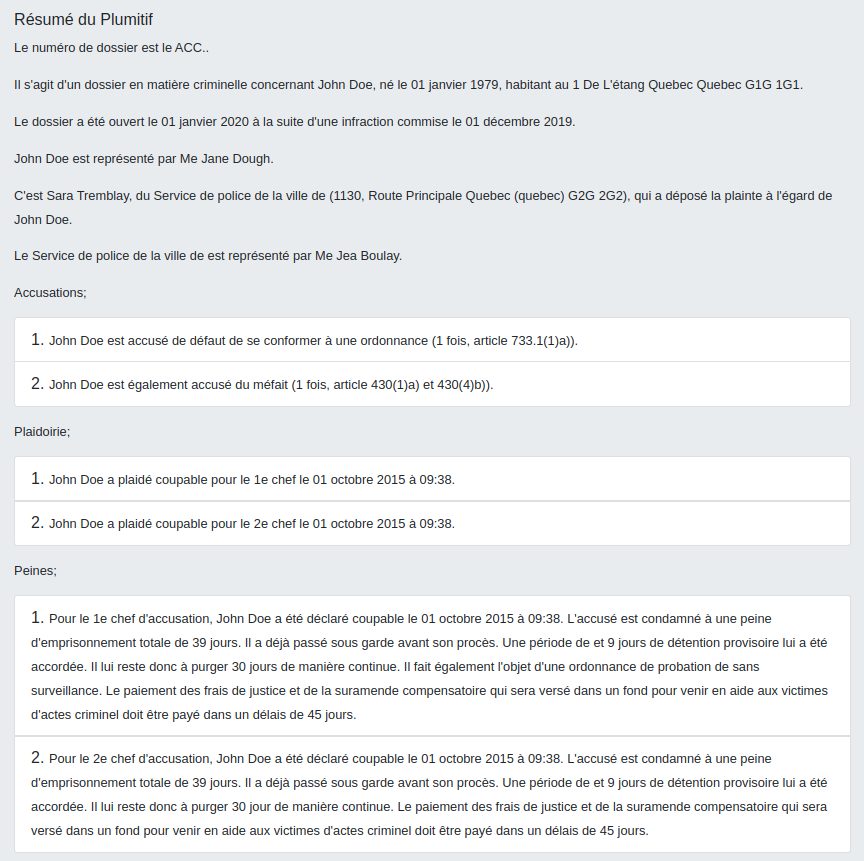}
    \captionlarge{Example of a complete generation using the \textit{plumitif} presented in the Figure \ref{fig:plumitif}.}
    \label{fig:plumitif_generation}
\end{figure}

\clearpage
\section{Web Application}
\label{sec:webapp}

\begin{figure}[h!]
    \centering
    \captionsetup{width=\textwidth}
    \includegraphics[width=\textwidth]{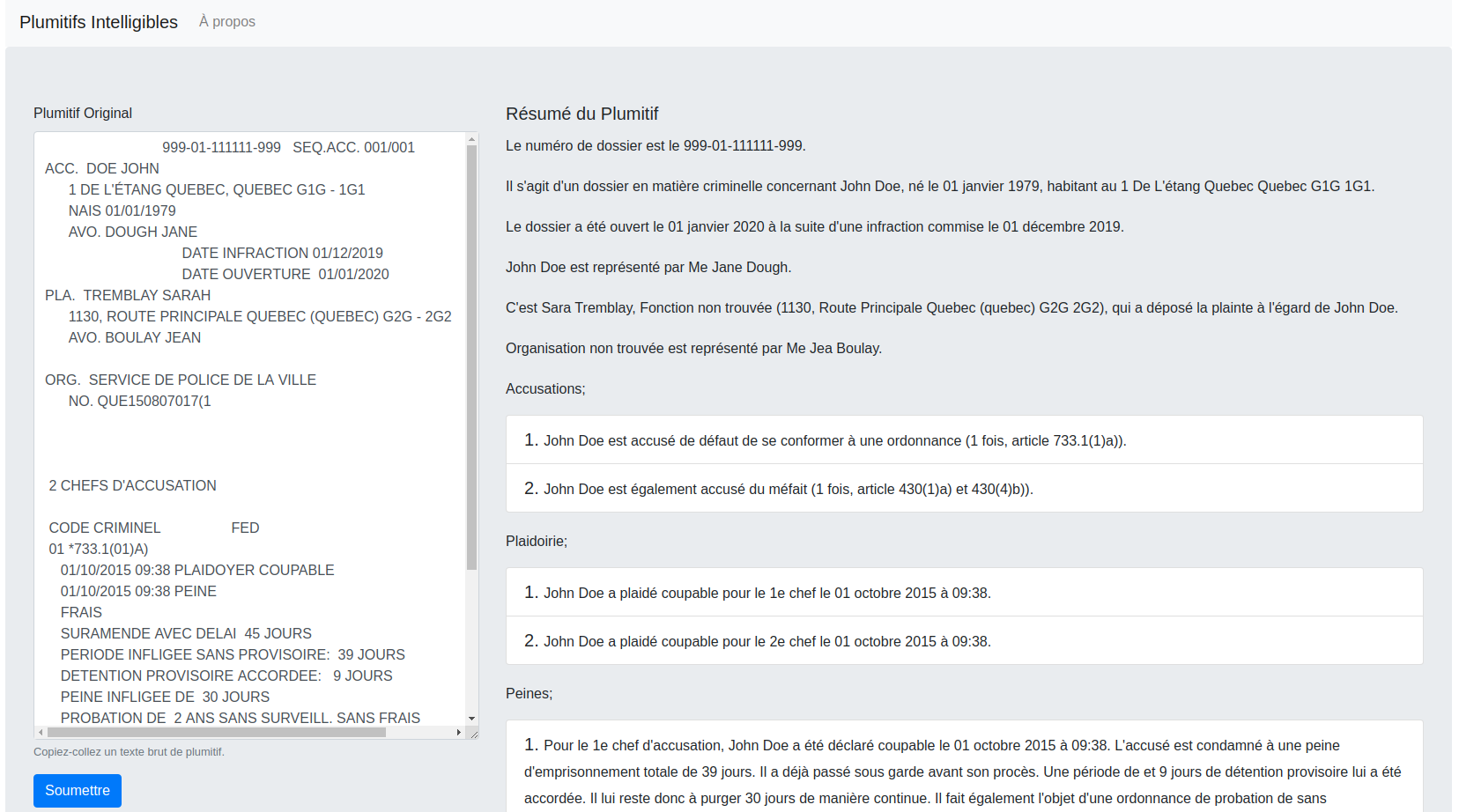}
    \caption{%
    Picture of the Web application.}%
    \label{fig:webapp}
\end{figure}

We developed a web application that is able to generate an intelligible summary from a raw \textit{plumitif}.
The workflow for litigants to obtain the \textit{plumitif}'s summary is fairly simple;
\begin{enumerate}
    \item Obtain the raw \textit{plumitif} from either the SOQUIJ website or physically at a district's court, as introduced in Section~\ref{subsec:ie}
    \item Copy and paste the raw \textit{plumitif} into the text area and submit the form. The summary will then be generated.
\end{enumerate}
This design is motivated, as discussed in Section~\ref{sec:ethic}, by a privacy concern, which refrains us from releasing these summaries in bulk for a lot or all available \textit{plumitifs}. It is important to say that there is not any \textit{plumitifs} available through this app, it is only ``translating'' \textit{plumitifs}  that citizens have on hand.
We present a picture of the web application in Figure~\ref{fig:webapp}.

\end{document}